\documentclass{article}
\pdfoutput=1

\usepackage{PRIMEarxiv}

\usepackage[utf8]{inputenc} 
\usepackage[T1]{fontenc}    
\usepackage{hyperref}       
\usepackage{url}            
\usepackage{booktabs}       
\usepackage{amsfonts}       
\usepackage{nicefrac}       
\usepackage{microtype}      
\usepackage{lipsum}
\usepackage{fancyhdr}       
\usepackage{graphicx}       
\graphicspath{{media/}}     
\usepackage{natbib}
\usepackage{booktabs}
\usepackage{enumerate}
\usepackage{blindtext}
\usepackage{multirow}
\usepackage{amsmath}
\usepackage{amsfonts}
\usepackage{float}
\usepackage{lipsum}

\title{
MDPO: Multi-Granularity Direct Preference Optimization for Mathematical Reasoning 
}

\author{
  Yunze Lin, \\
  School of Artificial Intelligence \\
  Beijing University of Posts and Telecommunications \\
  Beijing\\
  \texttt{linyunze@bupt.edu.cn} \\
}

\begin{document}
\maketitle

\begin{abstract}
Mathematical reasoning presents a significant challenge for Large Language Models (LLMs) as it requires ensuring the correctness of each reasoning step. Researchers have been strengthening the mathematical reasoning abilities of LLMs through supervised fine-tuning, but due to the inability to suppress incorrect outputs, illusions can easily arise.
Recently, Direct Preference Optimization (DPO) has been widely adopted for aligning human intent by using preference data to prevent LLMs from generating incorrect outputs. However, it has shown limited benefits in long-chain mathematical reasoning, mainly because DPO struggles to effectively capture the differences between accepted and rejected answers from preferences in long-chain data. The inconsistency between DPO training and LLMs' generation metrics also affects the effectiveness of suppressing incorrect outputs.
We propose the Multi-Granularity Direct Preference Optimization (MDPO) method, optimizing the mathematical reasoning of LLMs at three granularities: Solution2Solution, Inference2Inference, and Step2Step. Solution2Solution focuses on the correctness of entire long-chain reasoning; Inference2Inference concentrates on logical reasoning between steps; Step2Step corrects computational errors in steps, enhancing the computational capabilities of LLMs. Additionally, we unify the training objectives of the three granularities to align with the generation metrics.
We conducted experiments on the open-source models Qwen2 and Llama3, achieving improvements of 1.7\% and 0.9\% on the GSM8K dataset, and 2.3\% and 1.2\% on the MATH dataset, outperforming DPO and other DPO variant methods.
Furthermore, we also provide a pipeline for constructing MDPO training data that is simple and does not require manual annotation costs.
\end{abstract}

\section{Introduction}
Mathematical reasoning is considered a critical long-chain reasoning ability for large language models (LLMs) . This task is particularly challenging because it typically requires extensive chains of thought, potentially involving numerous reasoning steps. Any error in a reasoning step can lead to an incorrect final result.

Many studies have utilized additional math word problems (MWPs) data to conduct supervised fine-tuning (SFT) of LLMs for improving their mathematical reasoning abilities \cite{sft1,sft2,sft3}. However, models in the fine-tuning process often experience hallucinations, leading to performance saturation \cite{orpo}. On one hand, SFT struggles to provide fine-grained supervision signals; on the other hand, it is challenging for SFT methods to suppress the probability of undesirable outputs, making models more prone to errors in long-chain reasoning. Therefore, developing methods to provide fine-grained supervision signals and suppress the probability of undesirable outputs is crucial.

Recently, the direct preference optimization (DPO) alignment method uses preference data triplet \((x, y_w, y_l)\) to increase the probability of model outputting answers preferred by humans \(y_w\) while decreasing the probability of answers rejected by humans \( y_l\), and it is popular due to its simplicity \cite{dpo}. While DPO is effective in casual chat benchmarks, it struggles with long-chain mathematical reasoning.

Models using DPO perform poorly in distinguishing between correct and incorrect mathematical solutions, failing to accurately identify errors in incorrect solutions. For incorrect solutions, the model often generates the initial steps correctly, and using DPO may lower the probability of these correct steps. This indicates that LLMs fine-tuned with DPO cannot accurately pinpoint detailed errors in incorrect solutions, hindering the improvement of reasoning abilities.

Moreover, as highlighted in \cite{simpo}, during DPO training, for any triplet \((x, y_w, y_l)\), satisfying the reward ranking \(r(x, y_w) > r(x, y_l)\) does not necessarily imply satisfying the likelihood ranking \(p_{\theta}(y_w,x) > p_{\theta}(y_l,|x)\). In fact, only about 50\% of triplets meet this condition, which is due to the inconsistency between the fine-tuning and the ultimate downstream task.

In the daily learning of mathematics, a good teacher will clearly point out the errors when a student answers incorrectly, indicating whether the mistake is due to an error in derivation or calculation. At the same time, the teacher should provide the correct solution method and encourage students to reflect and learn from their mistakes.

In this paper, we propose Multi-granularity mathematical Direct Preference Optimization (MDPO) that provides models with supervision signals ranging from coarse to fine, unifying the representation of fine-tuning and final reasoning tasks to enhance the model's reasoning and computational abilities.

\begin{figure*}[htbp]
\centerline{\includegraphics{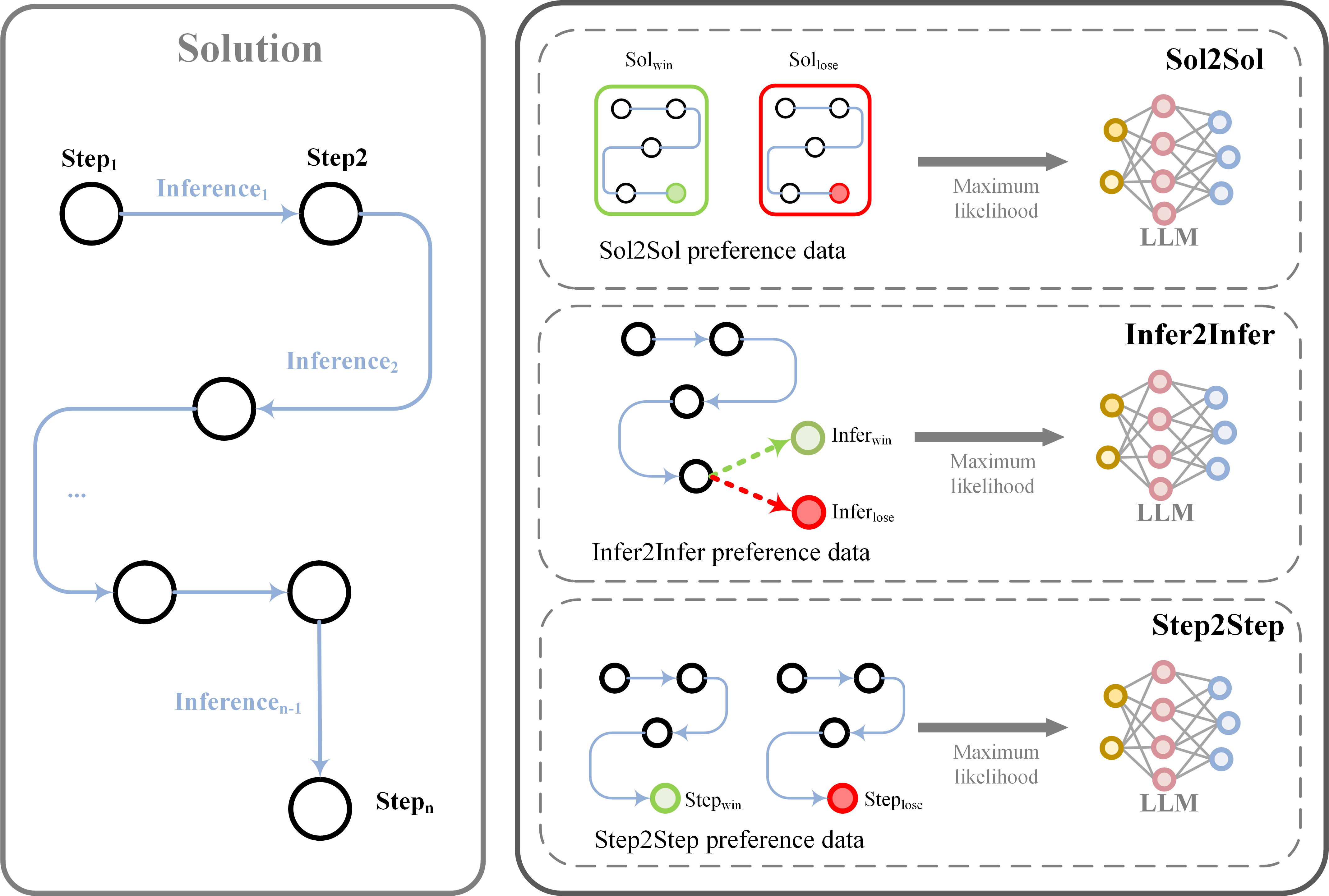}}
\caption{(Left) Given a math word problem, which, when fully reasoned through the solution chain, can be broken down into a series of \(\text{solutions} = \text{Step}_1, ..., \text{Step}_n\), we define the generation from \(\text{Step}_{k}\) to \(\text{Step}_{k+1}\) as one inference step. (Right) MDPO provides LLMs with three granularities of supervision signals: solution2solution (Sol2Sol), inference2inference (Infer2Infer), and step2step, optimizing the model using preferences for data. Sol2Sol constructs preferences for complete inference chains; Infer2Infer identifies faulty inferences in the chain and corrects them, while Step2Step focuses on identifying and correcting computational errors at a specific step in the inference process.}
\label{main}
\end{figure*}

Specifically, LLMs consider the entire chain of reasoning for solving MWPs as a \(solution\), composed of multiple reasoning steps, as shown in Fig.\ref{main} (left), \(solution = step_1, \dots, step_n\). We define the generation from \(step_k\) to \(step_{k+1}\) as one \(inference\).

As shown in Fig.\ref{main} (right), our method performs preference optimization at three granularities: Solution2Solution (Sol2Sol), Inference2Inference (Infer2Infer), and Step2Step. Sol2Sol provides the complete reasoning chain as the supervision signal, consistent with DPO, offering coarse-grained supervision. Infer2Infer locates the unreliable inference \(infer_{lose}\), corrects the model's reasoning process to obtain \(infer_{win}\), providing fine-grained supervision. Step2Step locates the step with computational errors, \(step_{lose}\), and corrects it to obtain \(step_{win}\), improving the model's computational capabilities. These task objectives are uniformly defined as given problem \(x\) and previous k steps \(step_{0\sim k}\), continue reasoning until generating the answer. Transforming the mathematical reasoning task into a text completion task ensures consistency between fine-tuning and ultimate downstream tasks.

We conducted experiments using two popular open-source models, Qwen2 \cite{qwen} and Llama3 \cite{llama}, on the GSM8K \cite{gsm} and MATH \cite{math} datasets. When using Qwen2-7B-Instruct, we achieved accuracy improvements of 1.7\% and 0.9\% on the GSM8K dataset, and 2.3\% and 1.2\% on the MATH dataset, surpassing DPO\cite{dpo} and its variant Step-DPO\cite{stepdpo}, demonstrating the potential of MDPO.

Additionally, we also provide a method for automatically constructing multi-granularity preference data pairs without the need for manual annotation, generating high-quality preference data.

\section{Related Work}
\subsection{Mathematical Reasoning}
With the increase in pre-training scale, LLMs have shown strong reasoning abilities. Reference \cite{cot} proposed the Chain-of-Thought, which allows LLMs to output a step-by-step reasoning process through artificially designed examples, not only improving the accuracy of solving math word problems but also enhancing the confidence in answers. Subsequently, some researchers attempted to further enhance the mathematical reasoning abilities of LLMs through strengthening the Chain-of-Thought \cite{autocot,zscot,plan,tot}. However, these efforts are often hindered by hallucinations and computational limitations of LLMs. 

Reference \cite{pal} introduced the Program-aided Language model (PAL), transforming the task of solving math word problems into a Python code generation task, ensuring the correctness of intermediate step calculations. However, this usually requires LLMs to have a relatively strong ability in code generation, as otherwise, more noise may be introduced during the code generation process. 

Additionally, many researchers utilized methods such as data augmentation to construct higher-quality MWPs data and use of new data for supervised fine-tuning of LLMs, significant improvements have been seen in the reasoning abilities of LLMs, yet the issue of hallucinations in LLMs remains unresolved \cite{sft1,sft2,sft3,sft4,sft5}. One potential reason is that supervised learning only increases the probability of generating correct answers and does not suppress the generation of incorrect answers. Therefore, it is essential to ensure both the generation of correct answers and the suppression of undesirable answers in improving the performance of LLMs.
\subsection{RLHF}
According to reinforcement learning from human feedback (RLHF), it is commonly used to train LLMs to align with human values and intentions, including suppressing the generation of harmful information and hallucinations \cite{rlhf1,rlhf2,rlhf3,rlhf4}. Recent studies have explored using reinforcement learning to reduce illusions in mathematical reasoning, involving training a reward model and then optimizing the policy model to maximize this reward. Reference \cite{prm} have proposed the Process Reward Model (PRM) to evaluate the quality of each reasoning step, thereby enhancing the mathematical reasoning abilities of the model. However, these methods not only require high-quality manually annotated data to train the reward model, but also involve complex and laborious training processes.

Recently, some researchers have explored simpler offline algorithms, one of which is Direct Preference Optimization (DPO) \cite{dpo}. DPO learns policy models directly from preference data by parameterizing the reward function in RLHF, eliminating the need for an explicit reward model. This method is simple, stable, widely used in practice, and has inspired various derivative methods \cite{kto,orpo,stepdpo,simpo}. One variant, Step-DPO, introduces supervised signals at the step level for LLMs to accurately locate errors, similar to our approach \cite{stepdpo}. However, its implicit reward is constructed by the logarithm of the likelihood ratio between responses from the current policy model and the Supervised Fine-Tuning (SFT) model, which does not align directly with the guided generation metrics, resulting in poor performance. Our method focuses on advancing LLMs in terms of computational capabilities.

Reference \cite{simpo} proposed the SimPO method, which directly uses average log-likelihood as a reward for preference learning, aligning training and reasoning, making it simple and efficient. Inspired by this research, we propose the MDPO method for mathematical long-chain reasoning, providing LLMs with multiple levels of supervised signals.
\section{Background: Simple Preference Optimization (SimPO)}
SimPO is one of the most popular preference optimization methods, which neither requires a reward model nor a reference model. It addresses the issue of discrepancies between the reward optimized during training and the generation metrics used during inference, outperforming DPO and other variant methods significantly. The core of the algorithm is to align the reward function in the preference optimization objective with the generation metric. Specifically, during generation, a policy model \(\pi_{\theta}\) is used to generate a sequence that approximates the maximization of the average log-likelihood, defined as follows:
\begin{align}
p_\theta(y \mid x) 
&= \frac{1}{|y|} \log \pi_\theta(y \mid x) \notag \\
&= \frac{1}{|y|} \sum_{i=1}^{|y|} \log \pi_\theta(y_i \mid x, y_{<i}). \
\end{align}

SimPO directly uses \(p_\theta\) from the equation \verb|(1)| as the reward, aligning it with the likelihood metric used for guided generation, while introducing a length normalization term to prevent the model from tending to generate longer but lower-quality sequences.

\begin{equation}
r_{\text{SimPO}}(x, y) = \frac{\beta}{|y|} \log \pi_\theta(y \mid x) = \frac{\beta}{|y|} \sum_{i=1}^{|y|} \log \pi_\theta(y_i \mid x, y_{<i}),
\end{equation}

Where \(\beta\) is a constant that controls the magnitude of the reward difference.

In addition, SimPO introduces a target reward margin term \(\gamma > 0\) to ensure that the reward \(r(x, y_w)\) of a winning response exceeds the reward \(r(x, y_l)\) of a losing response by at least \(\gamma \).
\begin{equation}
    p(y_w \succ y_l| x) = \sigma (r(x, y_w) - r(x, y_l) - \gamma)
\end{equation}

 The loss function is represented as follows:
\begin{align}
\mathcal{L}_{\text{SimPO}}(\pi_\theta) = 
    &-\mathbb{E}_{(x, y_w, y_l) \sim \mathcal{D}} 
    \Bigg[  \log \sigma \Bigg( 
    \frac{\beta}{|y_w|} \log \pi_\theta(y_w \mid x)  \\
    & - \frac{\beta}{|y_l|} \log \pi_\theta(y_l \mid x) \notag - \gamma 
    \Bigg) \Bigg]
\end{align}
\section{Method}

\subsection{MDPO}
Although DPO has proven effective in chat benchmarks, it brings only marginal improvements for long-chain reasoning tasks such as mathematical problems. This limitation arises because rejecting an entire undesirable answer in DPO may discard preceding correct reasoning steps, introducing significant noise and negatively impacting training. Analogous to how teachers correct students by pinpointing specific errors rather than dismissing entire answers, our proposed MDPO provides more detailed supervision from three granularities: Solution2Solution (Sol2Sol), Inference2Inference (Infer2Infer), and Step2Step. 

\textbf{Sol2Sol.} In solving mathematical word problems, the entire reasoning path from scratch to the final answer is referred to as a Solution, which aligns with the format in DPO. Although situations arise where rejecting the entire undesirable answer may have consequences, in order to ensure the model can generate a complete reasoning chain, we retain this practice. 


\textbf{Infer2Infer.} Each solution can be decomposed into a sequence of reasoning steps, \(\text{solution} = step_1, \dots, step_n\), where \(step_i\) is the i-th reasoning step. We define the generation from  \(step_k\) to \(step_{k+1}\) as an Inference. Given a prompt \(x\) and a series of initial correct reasoning steps \(step_{1 \sim k-1} = step_1, \dots , step_{k-1}\). Infer2Infer provides fine-grained supervision signals for the generation process \(infer_{k-1}\) from \(step_{k-1}\) to \(step_{k}\), maximizing the probability of generating the correct next reasoning step \(step_{k}^{win}\) , while minimizing the probability of generating the incorrect reasoning step \(step_{k}^{lose}\), enhancing the model's reasoning capabilities.

\textbf{Step2Step.} Since large models often face computational challenges during reasoning, leading to overall failure, we provide preference data for accepting and rejecting steps. Specifically, for a \(step_k^{lose}\) with computational error, we construct the correct calculation step \(step_k^{win}\) to rectify the model's computational errors and enhance its computational abilities.
\subsection{Objective}
To maintain consistency between fine-tuning and downstream tasks, in this paper, we convert mathematical word problem reasoning into the following form: Given a math word problem and the first k steps, the model is required to continue writing based on the problem and the first k steps to obtain the final answer. Adopting this format allows our multi-granularity optimization goals to align with the ultimate solving objectives.\\
\begin{itemize}
\item  For sol2sol, it can be considered as providing the problem and the first 0 steps, requiring the model to generate all reasoning steps, which aligns with the final solving of the math word problem. We use "Let’s think step by step." as the 0-th step to guide the model in reasoning; \\
\item  For infer2infer, it can be viewed as providing the problem and the first i-1 steps, asking the model to generate the i step and subsequent reasoning steps; \\
\item  For step2step, likewise, it can be seen as providing the problem and the first i-1 steps, demanding the model to generate the i-th step and subsequent reasoning paths.
\end{itemize}
\begin{align}
    &\mathcal{L}_{\text{MDPO}}(\pi_\theta) = \\
    &-\mathbb{E}_{(x,s_{0\sim k-1}, y_w, y_l) \sim \mathcal{D}} 
    \Bigg[\log \sigma \Bigg( 
    \frac{\beta}{|y_w|} \log \pi_\theta(y_w \mid (x,s_{0\sim k-1})) \notag \\
    &- \frac{\beta}{|y_l|} \log \pi_\theta(y_l \mid (x,s_{0\sim k-1})) \notag - \gamma 
    \Bigg) \Bigg]
\end{align}

Where \(x\) represents the math word problem to be solved, \(s_{0 \sim k-1}\) represents the first k solving steps from \(step_0\) to \(step_{k-1}\), \(y_w\) represents a series of correct solving steps from \(step_k\) to \(step_n\), and \(y_l\) represents a series of incorrect solving steps from \(step_k\) to \(step_n\).

\subsection{Datas Construction}\label{data}
\textbf{Sol2Sol. } In Sol2Sol part, we use LLMs to sequentially generate reasoning for questions and require the model to prepend "[Step i]" before each step. Subsequently, we sample k reasoning paths, and verify the generated paths based on labels in the dataset, adopting the complete reasoning path with the correct final answer as the preferred response, and the reasoning path with the incorrect final answer as the rejected response, consistent with simple DPO. Moreover, we select questions in the generated paths that have both correct and incorrect answers, which are more challenging for the model and more effective in enhancing the model's reasoning abilities.

\textbf{Infer2Infer. } In infer2infer part, we utilize the erroneous reasoning paths selected from the aforementioned problems as the foundation. These reasoning paths are then divided into steps and reorganized into multiple windows \(W = (w_0, \dots, w_i)\), where \(w_i = (step_0, step_1, step_2, \dots, step_i)\). Subsequently, we leverage LLMs to generate and sample \(k\) reasoning paths.

For each \(w_i\), we define its error rate as the number of erroneous reasoning paths divided by the total number of paths. Based on this, we define an unreliable step as a step where \(\text{error}(w_i) > \text{error}(w_{i-1})\).
We consider that \(step_i\) increases the reasoning error rate and negatively impacts the final reasoning process. As a result, the transition from \(step_{i-1}\) to \(step_i\) is regarded as an unreliable inference \(infer_{lose}\).

Next, we continue to generate using \(w_{i-1}\), sampling the reliable reasoning path with the final correct answer as \(infer_{win}\), to construct preference data pairs \((x||w_{i-1}, infer_{win}, infer_{lose})\), where \(||\) denotes the concatenation operation.


\textbf{Step2Step. } In the step2step part, we also used the selected problems mentioned above and added new problems involving complex calculations, which were constructed by replacing the original problems with more complex numbers. Subsequently, we used LLMs for reasoning, sampled the reasoning paths, and segmented the steps. By using prompts, we use GPT-4 to search for the first step, \(step_k^{lose}\), where a calculation error occurred, corrected it to obtain \(step_k^{win}\), and continued generating the final answers. Through answer verification, we ensured that the LLMs made correct modifications. We construct preference data \((x||step_{0\sim k-1},step_k^{win},step_k^{lose})\).


\section{Experiments}
\textbf{Network Architecture.} Our experiments used two currently popular open-source models, Qwen2 \cite{qwen} and Llama3 \cite{llama}. Due to limitations in computational resources, we utilized smaller-scale models, Qwen2-7B-Instruct and Meta-Llama-3-8B-Instruct. Our methodological choice of employing instruct models rather than base models for direct training stems from the common practice in reinforcement learning (RL) pipelines, where supervised fine-tuning (SFT) typically serves as a crucial warm-start initialization phase before proceeding with RL-based optimization.

\textbf{Datasets. } For evaluation, we used the widely adopted GSM8K\cite{gsm} and MATH\cite{math} datasets. Accuracy in these datasets serves as the evaluation metric. The MATH test set contains 5000 mathematical problems spanning 5 difficulty levels and 7 subjects, including algebra, counting and probability, geometry, intermediate algebra, number theory, prealgebra, and precalculus. Problems in GSM8K are generally easier than those in MATH. Additionally, we also use the GSM-HARD\cite{pal} dataset to examine the improvement in computational capabilities of LLMs by MDPO.

Our training dataset is based on the training sets of GSM8K and MATH, constructed following the method described in Section \ref{data}, comprising a total of 30,000 pairs of preference data.

\textbf{Implementation Details.} We perform MDPO on the models. For MDPO, we train the models for 8 epochs. The global batch size is set to 128, and the learning rate is set to 5e-7. The hyperparameter $\beta$ is set to 0.4 . We use the AdamW optimizer and a cosine learning rate scheduler, with the warmup ratio set to 0.1.
\begin{figure*}[htbp]
\centerline{\includegraphics{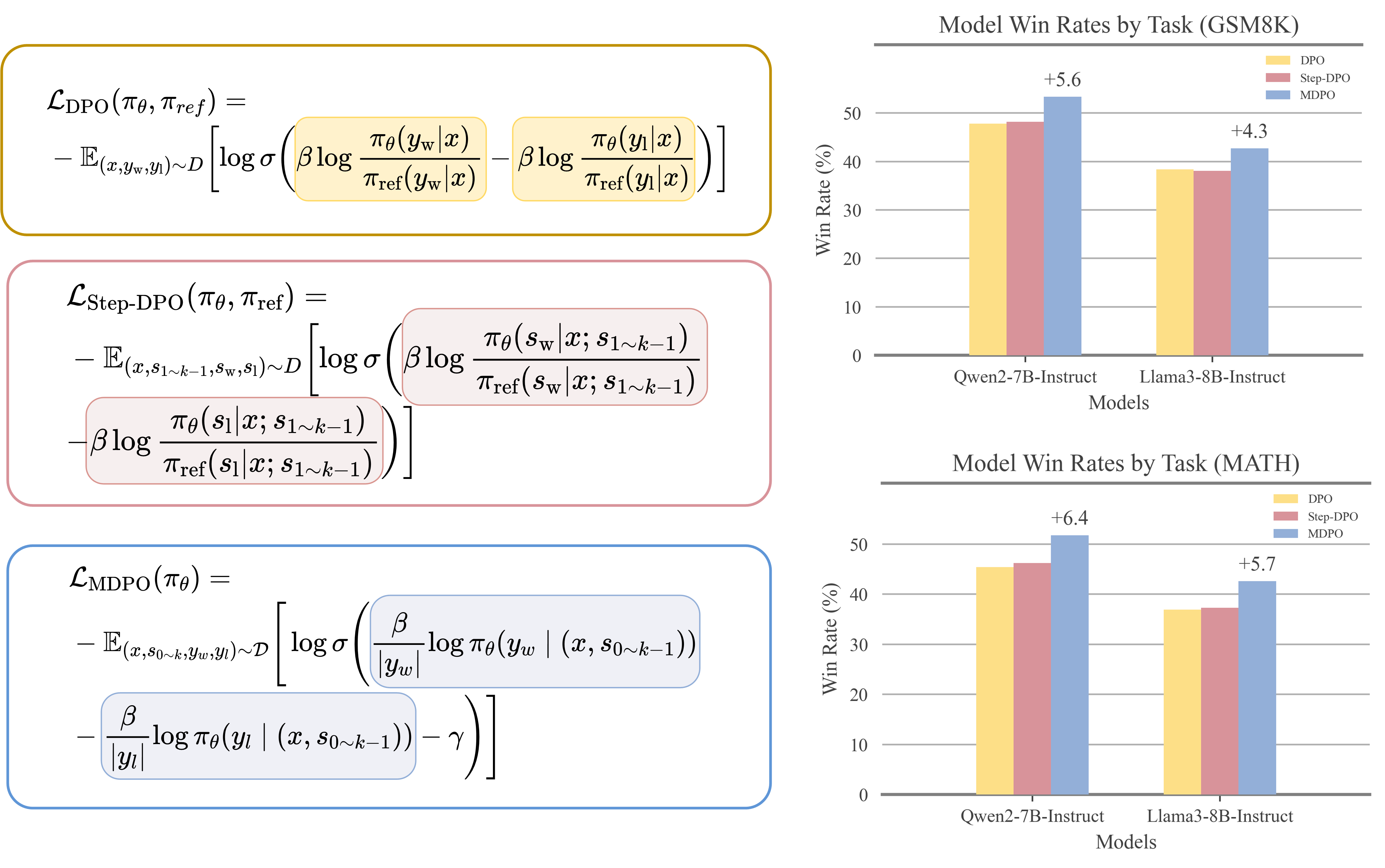}}
\caption{
MDPO has different training objectives from DPO and Step-DPO, mainly in the formulation of rewards and setting of reward margins. On the tasks of GSM8K and MATH, MDPO's consistency between training and downstream objectives has led to performance improvements, mainly reflected in the increased generation probability of accepting answers. The Win Rate indicates the proportion where \(p_{\theta}(y_w | x) > p_{\theta}(y_l | x)\).}

\label{logit}
\end{figure*}
\section{Results}
\subsection{Main Results}
\begin{table}[]
    \centering
    \caption{Math reasoning performance comparison on GSM8K and MATH across Qwen-7B-Instruct and Llama3-8B-Instruct.}
    \begin{tabular}{l|cc}
    \toprule
     Model& GSM8K(\%) & MATH(\%)\\
     \hline
     Qwen2-7B-Instruct \cite{qwen} & 81.7&54.2\\
     Qwen2-7B-Instruct + MDPO & 83.4 (+1.7)& 56.5 (+2.3) \\
     \hline
     Llama3-8B-Instruct \cite{llama} & 79.4 &30.0\\
     Llama3-8B-Instruct + MDPO & 80.3 (+0.9) & 31.2 (+1.2) \\
    \hline
    \end{tabular}
    
    \label{tab:main}
\end{table}

The results of MDPO on the Qwen2-7B-Instruct and Llama3-8B-Instruct models are shown in Table \ref{tab:main}. MDPO has shown significant improvements for both models, with an increase of 1.7\% in accuracy on the GSM8K dataset and an increase of 2.3\% in accuracy on the MATH dataset when applied to Qwen2-7B-Instruct. On the other hand, when applied to Llama3-8B-Instruct, Llama3-8B-Instruct + MDPO achieved a 0.9\% increase in accuracy on the GSM8K dataset and a 1.2\% increase in accuracy on the MATH dataset. This demonstrates that MDPO can effectively enhance the ability of LLMs in solving MWPs. 

Due to limitations in computational resources, we were only able to conduct experiments on small-scale LLMs. However, based on research on DPO and Step-DPO, methods proven effective on small-scale LLMs generally yield greater improvements on large-scale LLMs.
\subsection{Compare with other methods}
\begin{table}[]
    \centering
    \caption{Performance comparison of the MDPO method and other DPO methods on GSM8K and MATH datasets. }
    \begin{tabular}{l|cc}
    \toprule
     Model& GSM8K(\%)  & MATH(\%) \\
     \hline
     Qwen2-7B-Instruct \cite{qwen}& 81.7 &  54.2\\
     Qwen2-7B-Instruct + DPO \cite{dpo}& 81.9 & 54.6\\
     Qwen2-7B-Instruct + SimPO \cite{simpo}& 82.1 & 54.9\\
     Qwen2-7B-Instruct + Step-DPO \cite{stepdpo}& 82.1 & 55.1 \\
     Qwen2-7B-Instruct + MDPO & \textbf{83.4} & \textbf{56.5} \\
    \hline
    \end{tabular}

    \label{tab: compare}
\end{table}

We also compared our MDPO method with DPO and Step-DPO methods on the GSM8K and MATH datasets. The results are shown in Table \ref{tab: compare}, indicating that the benefits of DPO are limited and significantly less than those of MDPO. 

Furthermore, MDPO has shown improvements compared to Step-DPO on both datasets, with a greater enhancement on the MATH dataset. This is because MDPO not only optimizes the model in terms of reasoning ability but also provides supervisory signals for computational power, making it more effective on complex datasets.

We also conducted experiments using the original SimPO method. The empirical results demonstrate that SimPO achieves consistent performance improvements over standard DPO across both datasets, which can be attributed to its more direct reward signals. However, when compared with our proposed MDPO approach, SimPO exhibits inferior performance on both datasets, primarily due to its lack of fine-grained supervision signals at multiple levels. 
\subsection{Ablation Study}

\begin{table}[]
    \centering
    \caption{Ablation Study}
    \begin{tabular}{l|c}
    \toprule
     Model& GSM8K(\%) \\
     \hline
     Qwen2-7B-Instruct \cite{qwen}& 81.7\\
     Qwen2-7B-Instruct + sol2sol & 82.5 \\
     Qwen2-7B-Instruct + sol2sol + infer2infer & 83.2\\
     Qwen2-7B-Instruct + sol2sol + infer2infer + step2step & \textbf{83.4}\\ 
    \hline
    \end{tabular}

    \label{tab:ablation}
\end{table}
We conducted ablation experiments on MDPO, using Qwen2-7B-Instruct as the base model, and experimented on the GSM8K dataset. We sequentially added three granularities of preference data for fine-tuning the model, and the experimental results are shown in Table \ref{tab:ablation}. 

It can be seen that, compared to the base model, fine-tuning at each granularity has improved the final performance. It is noteworthy that the infer2infer part contributes the most to the model's performance improvement, as infer2infer provides finer supervisory signals to the model than sol2sol, making a greater contribution to enhancing the model's reasoning abilities. 

The step2step part aims to enhance the model's computational abilities. Since the operations in the GSM8K dataset are relatively simple, the contribution of step2step may not be fully reflected. For further experiments, please refer to Section \ref{compute}.

\subsection{Computation}\label{compute}

\begin{table}[]
    \centering
    \caption{In the performance comparison on the complex datasets GSM-HARD and MATH, where step2step indicates training using data constructed solely from the step2step section.}
    \begin{tabular}{l|cc}
    \toprule
     Model& GSM-HARD(\%)  & MATH(\%)\\
     \hline
     Qwen2-7B-Instruct \cite{qwen}& 42.1  & 54.2\\
     Qwen2-7B-Instruct + DPO \cite{dpo}& 42.6  & 54.6\\
     Qwen2-7B-Instruct + Step-DPO \cite{stepdpo}& 42.9   & 55.1\\
     Qwen2-7B-Instruct + step2step & 45.5  &55.9\\ 
    \hline
    \end{tabular}

    \label{tab:computation}
\end{table}

In the MDPO method, we introduced step2step to enhance the computational capabilities of the model. We used the GSM-HARD and MATH datasets to validate its effectiveness. In GSM-HARD, the numbers in the GSM8K test set were replaced with more complex digits, increasing the computational difficulty. The MATH dataset itself involves complex operations including fractions.

We conducted experiments with Qwen2-7B-Instruct and the step2step data. The experimental results are shown in Table \ref{tab:computation}. It can be seen that on GSM-HARD and MATH, the models fine-tuned using step2step data show a significant improvement (+3.4 and +1.7). Compared to the DPO and Step-DPO methods, there are clear advantages when dealing with computationally complex datasets. This demonstrates that incorporating fine-grained supervision signals for validating computations is essential.

\subsection{Training Objective}
In DPO and Step-DPO, the training objective does not align with the downstream task objective, which results in a higher probability of LLMs outputting rejected answers compared to accepted answers, contradicting the original intention of fine-tuning. 

We conducted experiments on the test set, and Fig. \ref{logit} shows the proportion of cases where LLMs output an accepted answer with higher probability after training with different methods. From the figure, it is evident that there is little difference between Step-DPO and DPO, as Step-DPO, while providing more detailed supervisory signals through modified optimization objectives, still fails to align the training objective with the downstream task. In contrast, our MDPO method aligns the reward function with the generation metric and unifies fine-tuning with the final downstream task, thereby enhancing LLMs' responsiveness to rewards. Experimental results demonstrate that the MDPO method significantly increases the probability of LLMs outputting accepted answers, effectively reducing the occurrence of rejected answers.

\section{Conclusion}
In this paper, we proposed the Multi-granularity Direct Preference Optimization (MDPO) method, which provides three granularities of supervision signals for LLMs: Solution2Solution, Inference2Inference, and Step2Step. The method optimizes the model using preferences for data. Solution2Solution ensures consistency between downstream tasks and fine-tuning, while Inference2Inference focuses on providing detailed problem-solving guidance for LLMs, accurately identifying logical errors in the reasoning process to improve the model's ability in long-chain reasoning. Step2step is dedicated to pinpointing computational errors in the inference process of LLMs, enhancing the model's foundational computational capabilities through data preferences.

Our method has been experimentally validated on two of the most popular open-source LLMs, Qwen2 and Llama3. It has shown significant improvements across multiple mathematical application datasets, generally outperforming DPO methods and Step-DPO methods. Specifically, on the MATH dataset, the MDPO method achieved a 2.3\% improvement, surpassing the Step-DPO method, which is widely recognized for its substantial gains in mathematical reasoning by 1.4\%.

Furthermore, our experiments on the challenging datasets GSM-HARD and MATH have demonstrated the effectiveness of step2step in enhancing the computational capabilities of LLMs within MDPO. In addition, in the experiment on target consistency, the necessity of aligning training objective with downstream tasks is showcased.

Due to computational resource limitations, our experiments have been conducted only on 7B-sized models. However, based on trends reported by other scholars\cite{dpo,simpo,stepdpo}, methods proven effective on small-scale models often lead to greater improvements when transplanted to large-scale models. In the future, we plan to enrich our experimental results by testing on a wider range of large-scale models.
\bibliographystyle{plainnat}  
\bibliography{main}  

\end{document}